\documentclass{llncs}

\usepackage[fleqn]{amsmath}
\usepackage[inline]{enumitem}
\usepackage{amssymb}
\usepackage{listings}
\usepackage{amsfonts}
\usepackage[normalem]{ulem}
\usepackage[backend=bibtex]{biblatex}
\usepackage{braket}
\usepackage{stmaryrd}
\usepackage{lingmacros}
\usepackage{enumitem}
\usepackage{linegoal}
\usepackage{xcolor}
\usepackage{cleveref}
\usepackage{bussproofs}
\usepackage{float}
\usepackage{tabularx}

\newcommand{\deppair}[2]{
  \begin{bmatrix}
    #1 \\
    #2 \\
  \end{bmatrix}
}

\newcommand{\event}{\textbf{event}}
\newcommand{\entity}{\textbf{entity}}
\newcommand{\type}{\textbf{type}}

\newcommand{\lenumsentence}[1]{
\enumsentence{
  \begin{minipage}[t]{\linegoal}
    \begin{enumerate}[label=\alph*.,leftmargin=*,ref=\theenums\alph*]
      #1
    \end{enumerate}
  \end{minipage}
}
}

\title{Handling Verb Phrase Anaphora with Dependent Types and Events}
\author{
  Daniyar Itegulov\inst{1, 2}
  \and
  Ekaterina Lebedeva\inst{1}
  \thanks{We thank Bruno Woltzenlogel Paleo and Florrie Verity for numerous helpful
    discussions and insightful remarks.}
}
\institute{
  Australian National University, Canberra, Australia \\
  \email{\{daniyar.itegulov, ekaterina.lebedeva\}@anu.edu.au}
  \and
  ITMO University, St. Petersburg, Russia \\
}
\begin{document}
\maketitle

\begin{abstract}
  This paper studies how dependent typed events can be used to treat
  verb phrase anaphora. We introduce a framework that extends Dependent Type
  Semantics (DTS) with a new atomic type for neo-Davidsonian events and an
  extended @-operator that can return new events that share properties of events
  referenced by verb phrase anaphora.

  The proposed framework, along with illustrative examples of its use, are
  presented after a brief overview of the necessary background and of the major
  challenges posed by verb phrase anaphora.
\end{abstract}

\section{Introduction}

Davidson~\cite{Davidson1967} observed that some verbs can
imply the existence of an ``action''. For example, a sentence ``John eats.''
represents an action of eating. This action can be anaphorically referred from a
following sentence (e.g.: ``The food is yummy.''). Hence, a framework of natural
language semantics should encompass the notion of action and a mechanism for
action reference. Davidson proposed to equip interpretations of verbs with an
additional argument for events. Then the sentence ``John
eats.'' is interpreted according to Davidson as $\exists e. eats(e, \textbf{j})$, instead of $eats(\textbf{j})$.

Parsons~\cite{Parsons1990} and Taylor~\cite{Taylor1986} argued that the approach
of event semantics captures the notion of adverbs better than approaches based
on higher-order predicates, such as~\cite{Verkuyl1972}, and is
easier to work with. For example, adverbial modifiers usually affect only the
event and not the entity, as the following example illustrates:

\lenumsentence{
  \item
    \label{butter}
    John buttered the toast slowly, deliberately, in the
    bathroom, with a knife, at midnight.
  \item
    \label{butter-dav-int}
    $\exists e. butter(e, \textbf{j}, \textbf{t}) \land slowly(e) \land
    deliberately(e) \land in(e, \textbf{b}) \land \exists k. with(e, k) \land
    at(e, \textbf{m})$
}

Sentence~(\ref{butter}) contains adverbs that modify the event of buttering the
toast. The corresponding interpretation is shown in~(\ref{butter-dav-int}).

Additionally to adverbial modifiers, Parsons~\cite{Parsons1990} described two
more reasons for introducing events as a new atomic type: perception verbs
and reference to events.

Parsons~\cite{Parsons1990}, furthermore, proposed a framework based on Davidson's
event theory, called neo-Davidsonian event semantics, that extends it as follows:

\begin{itemize}
  \item event participants are introduced via thematic roles
  \item state verbs, in addition to action verbs, are handled with an abstract variable
  \item two concepts, event holding and event culmination, are added
  \item events are decomposed into subevents
\end{itemize}

The differences between Davidsonian and neo-Davidsonian
approaches can be seen by comparing interpretations (\ref{butter-dav-int}) and
(\ref{butter-neodav-int}) of Sentence~(\ref{butter}).

\enumsentence{
  \label{butter-neodav-int}
  $\exists e. butter(e) \land agent(e, \textbf{j}) \land patient(e, \textbf{t})
  \land slowly(e) \land deliberately(e) \land in(e, \textbf{b}) \land \exists k.
  with(e, k) \land at(e, \textbf{m})$
}

This paper proposes a framework for solving verb phrase
anaphora (also known as verb phrase ellipsis) based on the neo-Davidsonian
event semantics and on dependent types; and to adapt the
existing techniques of handling the propositional anaphora to Dependent Type
Semantics framework. Dependent types are already used to express pronominal
anaphora in~\cite{Bekki2014}.

In Section 2, we briefly recall Dependent Type Semantics (DTS), which is a
theoretical foundation for our framework. In Section 3, we discuss major
problems of interpreting verb phrase anaphora. The main contribution of this
paper is presented in Section 4, which describes an extension of the Dependent
Type Semantics, and in Section 5, which discusses an application of subtyping in
the proposed framework.

\section{Recalling Dependent Type Semantics}

Dynamic type semantics (DTS) proposed in~\cite{Bekki2014} is a framework
of discourse semantics based on dependent type theory
(Martin-Löf~\cite{Martin1984}). DTS follows the constructive, proof-theoretic
approach to semantics established by Sundholm~\cite{Sundholm1986} who introduced
\emph{Sundholmian} semantics and by Ranta~\cite{Ranta1994} who introduced Type
Theoretical Grammar.

\begin{definition}[Dependent function]
  \label{dep-fun}
  For any $(s_1, s_2) \in \{(type, type), (type,\\
  kind), (kind, type), (kind, kind)\}, s \in \{type, kind\}$:
  \begin{figure}[H]
    \minipage{0.32\linewidth}
      \begin{prooftree}
        \alwaysNoLine
        \AxiomC{$A: s_1$}
        \AxiomC{$x: A$}
        \UnaryInfC{$\vdots$}
        \UnaryInfC{$B: s_2$}
        \alwaysSingleLine
        \BinaryInfC{$(x: A) \to B: s_2$}
      \end{prooftree}
    \endminipage\hfill
    \minipage{0.32\linewidth}
      \begin{prooftree}
        \alwaysNoLine
        \AxiomC{$A: s$}
        \AxiomC{$x: A$}
        \UnaryInfC{$\vdots$}
        \UnaryInfC{$M: B$}
        \alwaysSingleLine
        \BinaryInfC{$\lambda x. M: (x: A) \to B$}
      \end{prooftree}
    \endminipage\hfill
    \minipage{0.32\linewidth}
      \begin{prooftree}
        \AxiomC{$M: (x: A) \to B$}
        \AxiomC{$N: A$}
        \BinaryInfC{$MN : B[N/x]$}
      \end{prooftree}
    \endminipage\hfill
  \end{figure}
\end{definition}

\begin{definition}[Dependent pair]
  \label{dep-pair}
  For any $(s_1, s_2) \in \{(type, type), (type,\\
  kind), (kind, kind)\}$:
  \begin{figure}[H]
    \minipage{0.25\linewidth}
      \begin{prooftree}
        \alwaysNoLine
        \AxiomC{$A: s_1$}
        \AxiomC{$x: A$}
        \UnaryInfC{$\vdots$}
        \UnaryInfC{$B: s_2$}
        \alwaysSingleLine
        \BinaryInfC{$\deppair{x: A}{B}: s_2$}
      \end{prooftree}
    \endminipage\hfill
    \minipage{0.25\linewidth}
      \begin{prooftree}
        \AxiomC{$M: A$}
        \AxiomC{$N: B[M/x]$}
        \BinaryInfC{$(M, N): \deppair{x: A}{B}$}
      \end{prooftree}
    \endminipage\hfill
    \minipage{0.25\linewidth}
      \begin{prooftree}
        \AxiomC{$M: \deppair{x: A}{B}$}
        \UnaryInfC{$\pi_1M: A$}
      \end{prooftree}
    \endminipage\hfill
    \minipage{0.25\linewidth}
      \begin{prooftree}
        \AxiomC{$M: \deppair{x: A}{B}$}
        \UnaryInfC{$\pi_2M: B[\pi_1M/x]$}
      \end{prooftree}
    \endminipage\hfill
  \end{figure}
\end{definition}

DTS employs two kinds of dependent types (in addition to simply-typed lambda
calculus): \emph{dependent pair type} or $\Sigma$-type (notation $(x: A) \to
B(x)$) and \emph{dependent function type} or $\Pi$-type (notation $(x: A) \times
B(x)$). A dependent pair is a generalization of an ordinary pair.
By Curry-Howard correspondence between types and propositions, the type $(x:
A) \times B(x)$ corresponds to an existential quantified formula $\exists x^A.B$ and to an ordinary conjunction $A \land B$ when $x \not \in
fv(B)$.\footnote{$fv(x)$ denotes all free variables in $x$.}
A dependent function is a generalization of an ordinary function and the type
$(x: A) \to B$ corresponds to $\forall x^A. B$. Formal definitions are given through inference rules in~\cref{dep-fun,dep-pair}.

A comparison between the traditional notation for dependent types and the
notation used in DTS can be seen in~\cref{fig-dts-notation}.

\begin{figure}[h!]
  \center
  \begin{tabularx}{\textwidth}{ X | X | X }
    \hline
    & \centering $\Pi$-type & \centering \arraybackslash$\Sigma$-type \\
    \hline
    Initial notation & \centering $(\Pi x: A) B(x)$ & \centering \arraybackslash$(\Sigma x: A) B(x)$ \\
    DTS notation & \centering $(x: A) \to B(x)$ & \centering \arraybackslash$(x: A) \times B(x), \deppair{x: A}{B(x)}$ \\
    When $x \not \in fv(B)$ & \centering $A \to B$ & \centering \arraybackslash$\deppair{A}{B}$ \\
    \hline
  \end{tabularx}
  \caption{Notation in DTS}
  \label{fig-dts-notation}
\end{figure}

The main atomic type in DTS is $\textbf{entity}$, which represents all
entities in a discourse. With the employment of dependent type constructors, the entity type
can be combined with additional properties. For example,  $(\Sigma u: (\Sigma x:
\textbf{entity}) \times man(x)) \times enter(\pi_1(u))$ is a valid DTS interpretation of ``A
man entered''. Therefore, in contrast to traditional
approaches to semantical interpretation where entities are not distinguished by
their types, each entity has its own type in DTS.

In the traditional Montague model-theoretic semantics~\cite{Montague1974} a
proposition denotes a truth value (often defined as an $o$-type). However, DTS
does not follow this convention and instead the meaning of a sentence is
represented by a type. In dependent type theory, types are defined by the
inference rules, as shown in \cref{dep-fun,dep-pair}. The rules specify how a
dependent type (as a proposition) can be proved under a given context.
Thus, the meaning of a sentence in proof-theoretic semantics lies in its
\emph{verification condition} similar to the philosophy of language by
Dummett~\cite{Dummett1976}~\cite{Dummett1975} and Prawitz~\cite{Prawitz1980}.

To handle anaphora resolution, DTS distinguishes two kinds of propositions:
static and dynamic. A static proposition $P$ is called true if it is inhabited,
i.e. there exists a term of type $P$. A dynamic proposition is a function
mapping context-proof (a static proposition, representing the previous discourse) to a
static proposition.

@-operator is used to represent anaphora and presupposition
triggers. The operator takes the left context of dynamic propositions it is used
in. For example, Sentence~(\ref{self-love}) can be interpreted
as~(\ref{self-love-int}) in DTS. The @-operator in~(\ref{self-love-int}) takes a
context as an argument and tries to find a female entity (due to the
interpretation of the word ``herself'') in the context passed to it.

Different @-operators can have different types (since context passed to them can
vary in its type). They are distinguished with a numerical subscript. Since
different @-operators can take differently typed contexts, the operators should
have different types as well. That is why contexts' types of @-operators are
distinguished with a numerical subscript. The full type of $@_i$-operator can
look like this: $@_i: \gamma_i \to \textbf{entity}$.

\lenumsentence{
  \item
    \label{self-love}
    Mary loves herself.
  \item
    \label{self-love-int}
    $\lambda c. loves(\textbf{m}, \pi_1(@_1c: \deppair{x:
      \textbf{entity}}{female(x)}))$
}

\section{Verb Phrase Anaphora} \label{verb-phrase-anaphora}

Verb phrase anaphora~\cite{Prust1994} are anaphora with an
intentional omission of part of a full-fledged verb phrase when the ellipsed
part can be implicitly derived from the context. For example, verb phrase anaphora can be observed in~(\ref{john-left}) and~(\ref{mary-did-too}):

\lenumsentence{
  \item \label{john-left} \text{John left before Mary did.}
  \item \label{mary-did-too} \text{John left. Mary did too.}
}

In~(\ref{john-left}), the word ``did'' refers to an action John did
before Mary. In~(\ref{mary-did-too}), the ``did too'' clause refers to an
action which John and Mary both did. These sentences can be interpreted in event semantics as the following logical expressions:

\lenumsentence{
  \item $\exists e. agent(e, \textbf{j}) \land left(e) \land \exists e'.
    agent(e', \textbf{m}) \land left(e') \land before(e, e')$
  \item $\exists e. agent(e, \textbf{j}) \land left(e) \land \neg \exists e'.
    agent(e', \textbf{m}) \land left(e')$
}

Furthermore, an anaphoric verb phrase can ``inherit'' some properties from its
referent. Consider Example~(\ref{cake}) where ``did too'' not only refers to the event of eating
performed by John, but also to properties such as ``quietly'' and ``last
night''. Expression (\ref{cake-int}) is the interpretation of this sentence in event semantics.

\lenumsentence{
  \item
    \label{cake}
    John quietly ate the cake last night. Mary did too.
  \item
    \label{cake-int}
     $\exists e. (agent(e, \textbf{j}) \land patient(e, \textbf{c}) \land ate(e)
     \land quietly(e) \land at(e, \textbf{ln})) \land\\
     \exists e'. (agent(e',\textbf{m}) \land patient(e', \textbf{c}) \land
     ate(e') \land quietly(e') \land at(e', \textbf{ln}))$
}

A verb phrase anaphora may have an additional property that can ease the
choice of a correct anaphoric referent-event from the context. This phenomenon is
exemplified in~(\ref{correct-pasta}), where it is explicit that ``too'' refers to an
action connected with eating. 

\eenumsentence{
  \item John ate pasta and did not feel well. Mary ate too, but nothing
    happened to her. \label{correct-pasta}
}

An ambiguity between strict and sloppy identity readings of verb phrase anaphora
described by Prust~\cite{Prust1994} is another intriguing phenomenon.
Example~(\ref{hat-all}) illustrates this:

\eenumsentence{
  \label{hat-all}
  \item \label{hat}
    John likes his hat. Fred does too.
  \item \label{hat-int-1}
    $\exists x. hat(x) \land owner(x, \textbf{j}) \land \exists e. like(e)
    \land agent(e, \textbf{j}) \land patient(e, x) \land\\
    \exists e'. like(e') \land agent(e', \textbf{f}) \land patient(e', x)$
  \item \label{hat-int-2}
    $\exists x. hat(x) \land owner(x, \textbf{j}) \land \exists e. like(e)
    \land agent(e, \textbf{j}) \land patient(e, x) \land\\
    \exists y. hat(y) \land owner(y, \textbf{f}) \land \exists e'. like(e')
    \land agent(e', \textbf{f}) \land patient(e', y)$
}

The anaphoric clause in the second sentence of~(\ref{hat}) can be interpreted as ``Fred likes John's hat''
(the sloppy identity interpretation~(\ref{hat-int-1})) or as ``Fred likes Fred's
hat'' (the strict interpretation~(\ref{hat-int-2})). A desirable framework should
be able to provide both interpretations.

\section{Events with Dependent Types}

To tackle phenomena discussed in Section~\ref{verb-phrase-anaphora}, we propose
to extend DTS with a new atomic type $\textbf{event}$ for interpreting events.
Then, given its left contect $c$, DTS's @-operator can be employed for
retrieving a variable of type $\textbf{event}$ analogously to its original use
for retrieving a referent of type $\textbf{entity}$.

As was shown by Parsons~\cite{Parsons1990}, event semantics can be employed to
represent propositional anaphora. An example of propositional anaphora is shown
in~(\ref{cruel-love}), where ``this'' refers to the whole proposition expressed
in the first sentence. Formula~(\ref{cruel-love-int}) is an interpretation
of~(\ref{cruel-love}).

\lenumsentence{
  \item \label{cruel-love} John loved Mary. But Mary did not believe this.
  \item \label{cruel-love-int} $\exists e. agent(e, \textbf{j}) \land patient(e,
\textbf{m}) \land loved(e) \land \exists e'. believed(e') \land agent(e',
\textbf{m}) \land patient(e', e)$
}

Dependent typed events allow us to handle more complex types of
propositional anaphora. Similar to entities, events can have various properties
provided by their description. Assume the following three sentences appear in
the same discourse, possibly remotely from each other, but with preservation of
the order:

\lenumsentence{
  \label{canberra-flood}
  \item Canberra was hit by a flood on Sunday. \label{flood}
  \item The fair was held in London. \label{fair}
  \item What happened in Canberra is surprising. \label{surprising}
}

Here the anaphoric clause in Sentence~(\ref{surprising}) refers to an event discussed earlier. There are however (at least) two potential events for the reference: one given by~(\ref{flood}) and another given by~(\ref{fair}). Since the anaphoric clause in~(\ref{surprising}) specifies that it refers to an event happened  in Canberra, the anaphora disambiguates to the event in~(\ref{flood}).

The interpretation of verb phrase anaphora is more challenging, however, than the
interpretation of propositional anaphora: an anaphoric clause in a verb
phrase anaphora usually talks about a new event that inherits properties of
another event. For example, ``John left. Bob did too.'' conveys two events:
one is about John leaving and the second one is about Bob leaving. In cases of
pronominal and propositional anaphora, however, there is just a reference to an
entity or an event in the context. For example in ``John walks. He is slow.'',
pronoun ``he'' in the second sentence just refers to the entity ``John'' from
the first sentence.

To handle verb phrase anaphora correctly, it is not enough to
just fetch a referenced variable from the left context; instead a new variable of
type $\textbf{event}$ should be introduced. This new variable \emph{copies}
properties from the referred event. Furthermore, the agent of the referred event
should be changed to the current agent in the new event. This can be seen in
interpretation~(\ref{cake-int}) of~(\ref{cake}), where the agent John is replaced
with Mary.

Although $@_i$-operator has type $\gamma_i \to \textbf{entity}$ in DTS for handling
pronominal anaphora, according to DTS syntax for raw terms the operator can be
of any type. We therefore suggest a new type of @-operator that
guarantees that the returned event has a proper agent, necessary for
interpreting the verb phrase anaphora:

\enumsentence{
  \label{at-op-type}
$@_i: (c: \gamma_i) \to (x: \textbf{entity}) \to \deppair{e: \textbf{event}}{agent(e, x)}$ \label{@i}
}

Formula (\ref{first-dts-example}) is an interpretation of discourse~(\ref{mary-did-too}).
The $@_1$-operator in~(\ref{second-dts-example}) is applied
to its left context $c$ (of type~$\gamma_0$) and an entity, and returns a \emph{new}
event of type $\textbf{event}$ with the same properties (apart from the agent
property) of the referenced event. Crucially, the event returned by $@_1$-operator
in~(\ref{second-dts-example}) is not an event that was in the context previously. It is a
new event with the same properties (e.g. location, time) as a referenced event from
the context, but with a replaced agent.

\lenumsentence{
  \item
    \label{first-dts-example}
    $\lambda c. \deppair{e: \textbf{event}}{\deppair{left(e)}{agent(e, \textbf{j}})}$
  \item
    \label{second-dts-example}
    $\lambda c. (@_1c: (x: \textbf{entity}) \to \deppair{e:
      \textbf{event}}{agent(e, x)})(\textbf{m})$
}

Note that the entity accepted by the @-operator defined in~(\ref{@i}) is the
agent in the new event. For instance, in Example~(\ref{cake}), the interpretation of
``did too'' using~(\ref{@i}) would have the agent ``John'' of the referenced
event replaced by ``Mary'', but the patient (i.e. the cake) would remain. On the
other hand, there exist cases of verb phrase anaphora where the patient in the
referenced event should be replaced. This usually depends on the voice (active
or passive) of an anaphoric clause, as can be seen from examples in~(\ref{love-all}).

\enumsentence{
  \label{love-all}
  \begin{minipage}[t]{\linegoal}
  \begin{enumerate*}[label=\alph*.,ref=\theenums\alph*]
    \item
      \label{love-passive}
      Mary is loved by John. So is Ann.
      \\
    \item
      \label{love-active}
      John loves Mary. So does Bob.
      \\
    \item
      \label{love-passive-int}
      $
      \deppair
        {u:\deppair
            {e: \textbf{event}}
            {\deppair{agent(e, \textbf{j})}{\deppair{patient(e, \textbf{m})}{loved(e)}}}
        }
        {\deppair
          {e': \textbf{event}}
          {\deppair{agent(e', \textbf{j})}{\deppair{patient(e', \textbf{a})}{loved(e')}}}
        }
      $
      \quad
    \item
      \label{love-active-int}
      $
      \deppair
        {u:\deppair
            {e: \textbf{event}}
            {\deppair{agent(e, \textbf{j})}{\deppair{patient(e, \textbf{m})}{loved(e)}}}
        }
        {\deppair
          {e': \textbf{event}}
          {\deppair{agent(e', \textbf{b})}{\deppair{patient(e'''', \textbf{m})}{loved(e')}}}
        }
      $
  \end{enumerate*}
  \end{minipage}
}

The first sentences in (\ref{love-passive}) and (\ref{love-active}) have
the same semantics and hence the interpretations given to them in
(\ref{love-passive-int}) and (\ref{love-active-int}) coincide.
However, despite both second sentences are written in the same voice as their
first sentences, the second sentences are interpreted differently. Naturally,
the second sentence in (\ref{love-passive}) means ``Ann is loved by John'',
while the second sentence in (\ref{love-active}) means ``Bob loves Mary''. Note
that they have replaced different participants of the first sentences: in
(\ref{love-passive}) Mary (patient) was replaced by Ann and in
(\ref{love-active}) John (agent) was replaced by Bob.

Furthermore, an
interpretation of a sentence may require both the agent and the patient to be
replaced, as for example in the sloppy reading of~(\ref{hat-int-2}). These
possible cases of anaphora resolution can be tackled with the
judgements~(\ref{anaphora-operators}) assuming they occur in a global context
$\mathcal{K}$.

Another important notion in DTS is the felicity condition. The anaphora
resolution for $@_i$ operator is launched by type checking of the following
judgement: $\mathcal{K}, \gamma_i: type \vdash @_i: \gamma_i \to type$. It means
that the semantical interpretation of a sentence must be of the sort
\texttt{type} assuming that the left context is of type $\gamma_i$.
A requirement of a success of the launching the type checker is called
\emph{felicity condition}.

In order to preserve the original DTS invariants, we should show how felicity
condition is being fulfilled in the extended DTS. An example of a
felicity-judgement generated by verb phrase anaphora is shown in example
(\ref{felicity-condition-new}). It is different from the felicity condition from
original DTS notion since the new @-operator has a new type as shown in Equation~\ref{at-op-type}.

\enumsentence{
  \label{felicity-condition-new}
  $\mathcal{K}, \gamma_i: type \vdash @_i: \gamma_i \to (x: \textbf{entity}) \to
  \deppair{e: \textbf{event}}{agent(e, x)}$
}

Let us assume that the global context $\mathcal{K}$ contains the judgements from
(\ref{anaphora-operators}). Then one should be able to type check judgements
generated by the verb phrase anaphora.

\lenumsentence{
  \label{anaphora-operators}
  \item
    $\begin{aligned}[t]
      replaceA:\ &(p: \entity \to (e: \event) \to \type) \to\\
      & (original: \entity) \to (new: \entity) \to \\
      &(u: \deppair{e': \event}{p\ original\ e'}) \to
      (v: \deppair{e'': \event}{p\ new\ e''})
    \end{aligned}$
  \item
    $\begin{aligned}[t]
      replaceP:\ &(p: \entity \to (e: \event) \to \type) \to\\
      &(original: \entity) \to
      (new: \entity) \to \\
      &(u: \deppair{e': \event}{p\ original\ e'}) \to
      (v: \deppair{e'': \event}{p\ new\ e''})
    \end{aligned}$
  \item
    $\begin{aligned}[t]
      replaceAP:\ &(p: \entity \to \entity \to (e: \event) \to \type) \to \\
      &(oagent: \entity) \to
      (nagent: \entity) \to \\
      &(opatient: \entity) \to
      (npatient: \entity) \to \\
      &(u: \deppair{e': \event}{p\ oagent\ opatient\ e'}) \to
      (v: \deppair{e'': \event}{p\ nagent\ npatient\ e''})
    \end{aligned}$
  \item $j: \textbf{entity}$
}

Functions $replaceA$, $replaceP$, $replaceAP$ construct a new event $v$ from an
existing event $u$. To express the inheritance of properties and the change of
the agent in $replaceA$ (or patient in $replaceP$), properties are expressed as
a function that accepts two arguments: an agent-entity (or patient-entity in
$replaceP$) and an event; and returns a logical expression describing the event
using the entity. Function $replaceAP$ accounts for cases where both an agent
and a patient are replaced.

We can now construct term $@_1$ of type~(\ref{@i}), to fulfill the felicity
condition of form (\ref{felicity-condition-new}), as shown
in~(\ref{anaphora-resolution}):

\enumsentence{
  \label{anaphora-resolution}
  $\mathcal{K}, \gamma_0: type \vdash \begin{aligned}[t]
    &@_1: \gamma_0 \to (x: \textbf{entity}) \to \deppair{e: \textbf{event}}{agent(e, x)} =\\
    &\lambda c. \lambda x. replaceA\ (\lambda y. \lambda e. \deppair{left(e)}{agent(e, y)})\ \textbf{j}\ x\ \pi_1\pi_2(c)
  \end{aligned}$
}

A substitution of $@_1$ in~(\ref{second-dts-example}) with its term defined
in~(\ref{anaphora-resolution}) leads to the following semantical interpretation:

\enumsentence{
  \label{mary-did-too-int}
  $\deppair{e'': \textbf{event}}{\deppair{left(e'')}{agent(e'', \textbf{m})}}$
}

Since anaphora in DTS is resolved using the type checking procedure, the verb
phrase anaphora, just like the pronominal anaphora, can be resolved in various
ways. A type checking algorithm can find different terms which conform to the
specified (by felicity condition) type. For example, in order to handle the
ambiguity between strict and sloppy identity readings, which were discussed in
Example~(\ref{hat-all}), our framework can provide both possible
interpretations for Sentence~(\ref{hat}). Term~(\ref{hat-dtsi-int}) shows a
generic interpretation of~(\ref{hat}) in the proposed framework.

\enumsentence{
  \label{hat-dtsi-int}
    $
    \lambda c.
    \deppair
    {u:
      \deppair
      {v:
        \deppair
        {x: \textbf{entity}}
        {\deppair{hat(x)}{owner(x, \textbf{j})}}
      }
      {
        \deppair
        {e: \textbf{event}}
        {\deppair{like(e)}{\deppair{agent(e, \textbf{j})}{patient(e, \pi_1(v))}}}
      }
    }
    {
      @_1(c, u)\textbf{f}: \deppair{e': \textbf{event}}{agent(e', \textbf{f})}
    }
    $
}

In~(\ref{hat-dtsi-int}), ``Fred does too.'' is interpreted as the term $@_1(c,
u)\textbf{f}$, where $u$ stands for the interpretation of the preceeding
sentence ``John likes his hat.''. Recall from Example~(\ref{hat-all}) that the
latter sentence has an ambiguous meaning. (\ref{hat-anaphora}) defines two
alternative terms for $@_0$, one for each of the possible meanings. Note that
the type of these terms for $@_0$ conforms with the felicity condition.

\lenumsentence{
  \label{hat-anaphora}
\item
  $\begin{aligned}[t]
    \mathcal{K} \vdash
    &@_1: \gamma_0 \to (x: \textbf{entity}) \to \deppair{e': \textbf{event}}{agent(e', x)} =\\
    &\lambda c. \lambda f. replaceA\ (\lambda y.\lambda
    e.\deppair{like(e)}{\deppair{agent(e, y)}{patient(e, x)}})\ \textbf{j}\ f\
    \pi_1\pi_2\pi_2(c)
  \end{aligned}$
\item \label{hat-anaphora-2}
  $\begin{aligned}[t]
    \mathcal{K} \vdash &@_1: \gamma_0 \to (x: \textbf{entity}) \to \deppair{e':
      \textbf{event}}{agent(e', x)} =\\
    &\textbf{let}\ p = \lambda y. \lambda z. \lambda e.
      \deppair{like(e)}{\deppair{agent(e, y)}{patient(e, z)}}\\
    &\textbf{in}\ \lambda c. \lambda f.\deppair{u: \deppair{y:
      \textbf{entity}}{hat(y) \land owner(y, f)}}{replaceAP\ p\ \textbf{j}\ f\
      \pi_1\pi_1\pi_2(c)\ \pi_1(u)\ \pi_1\pi_2\pi_2(c)}
  \end{aligned}$
}

Both terms are valid substitutions for $@_1$-operator in~(\ref{hat-dtsi-int}) and
they represent strict and sloppy anaphora readings respectively.
In~(\ref{hat-anaphora-2}) let-in structure is used only as a syntactical sugar
for readability and is not actually a part of DTS term syntax.

\section{Subtyping}
Equation in~(\ref{anaphora-resolution}) (i.e. an anaphora resolution
solution: a proof of existence of a term with the required type under the global
context $\mathcal{K}$) is not sound: the type of the right side of the equation is $$\gamma_0 \to (x:
\textbf{entity}) \to \deppair{e'':
  \textbf{entity}}{\deppair{left(e'')}{agent(e'', x)}}$$ while the type required
by the left side is $$\gamma_0 \to (x: \textbf{entity}) \to \deppair{e:
  \textbf{entity}}{agent(e, x)}$$ The former type is more specific
than the latter type because it has an additional property ``left''.

This is not a problem, as events have a natural subtyping relationship between them. As
described in~\cite{Luo2017}, an event whose agent is $a$ and patient is $p$,
is an event with agent $a$. Despite a different theory underneath, the
techniques described there can be reused for subtyping events in DTS. This leads
to the following subtyping relations in event semantics:

\lenumsentence{ \label{subtyping}
\item $Evt_{AP}(a, p) <: Evt_A(a) <: Event \longleftrightarrow\\
      \deppair{e:
      \textbf{event}}{\deppair{agent(e, a)}{patient(e, p)}} <: \deppair{e:
      \textbf{event}}{agent(e, a)} <: \deppair{e: \textbf{event}}{()}$
  \item $Evt_{AP}(a, p) <: Evt_P(p) <: Event \longleftrightarrow\\
      \deppair{e:
      \textbf{event}}{\deppair{agent(e, a)}{patient(e, p)}} <: \deppair{e:
      \textbf{event}}{patient(e, p)} <: \deppair{e: \textbf{event}}{()}$
}

Subtyping relations of events can also depend on other properties
(e.g. a loud event performed by John is also an event performed by John). We
employ Luo's notation to define a new type $Event_{NA}(n, a)$, which is the type
of events with agent $a$ and nature $n$. Nature is a main predicate for each
event in neo-Davidsonian semantics (e.g. ``left(e)'' for an event of leaving,
``ate(e)'' for an event of eating).

The following transformation shows how the dependent event types in DTS notation
from~(\ref{anaphora-resolution}) can be converted into dependent event types
in Luo notation:

\lenumsentence{
  \item
    $\deppair{e'': \textbf{event}}{\deppair{left(e'')}{agent(e'', x)}} \longleftrightarrow
    e'': Event_{DA}(left, x)
    $
  \item
    $\deppair{e: \textbf{entity}}{agent(e, x)} \longleftrightarrow
    e: Event_{A}(x)
    $
}

A subtyping relationship between these types can be
constructed (assuming the appropriate subtyping rules have been added along with
type $Event_{DA}(d, a)$).

\begin{prooftree}
  \AxiomC{$left: Description$}
  \AxiomC{$x: Agent$}
  \BinaryInfC{$Event_{DA}(left, x) <: Event_A(x)$}
\end{prooftree}

The discussed subtyping relationship allows us to obtain~(\ref{anaphora-resolution}).




\section{Conclusion}
This paper introduces dependent event types for resolving verb phrase anaphora with
DTS as the underlying framework. To tackle verb phrase anaphora, we extend DTS's
@-operator, which was originally introduced for
handling pronominal anaphora. The paper also adresses strict and sloppy readings of verb phrase
anaphora and shows that each of them can be achieved solely by
manipulating the interpretation of the @-operator. The previous approaches to
handling the propositional anaphora were also adapted to DTS framework.

Techniques described in this paper could be applied to handle other cases of
anaphora, such as adjectival anaphora, modal and ``do so'' anaphora. Another
interesting topic would be to study specific behaviours of various thematic
roles, such as experiencer, theme and source.

\printbibliography

\end{document}